\documentclass[conference]{IEEEtran}
\IEEEoverridecommandlockouts
\usepackage{cite}
\usepackage{url}
\usepackage{amsmath,amssymb,amsfonts}
\usepackage{algorithmic}
\usepackage{graphicx}
\usepackage{textcomp}
\usepackage{xcolor}
\usepackage{float} 

\def\BibTeX{{\rm B\kern-.05em{\sc i\kern-.025em b}\kern-.08em
    T\kern-.1667em\lower.7ex\hbox{E}\kern-.125emX}}
\usepackage{caption}
\begin{document}

\title{DisasterQA: A Benchmark for Assessing the performance of LLMs in Disaster Response\\
}

\author{\IEEEauthorblockN{Rajat Rawat}
Fremont, USA \\
rajat.s.rawat@gmail.com}

\maketitle

\begin{abstract}
Disasters can result in the deaths of many, making quick response times vital. Large Language Models (LLMs) have emerged as valuable in the field. LLMs can be used to process vast amounts of textual information quickly providing situational context during a disaster. However, the question remains whether LLMs should be used for advice and decision making in a disaster. To evaluate the capabilities of LLMs in disaster response knowledge, we introduce a benchmark: DisasterQA created from six online sources. The benchmark covers a wide range of disaster response topics. We evaluated five LLMs each with four different prompting methods on our benchmark, measuring both accuracy and confidence levels through Logprobs. The results indicate that LLMs require improvement on disaster response knowledge. We hope that this benchmark pushes forth further development of LLMs in disaster response, ultimately enabling these models to work alongside emergency managers in disasters. 
\end{abstract}

\begin{IEEEkeywords}
LLM, disaster response, emergency management, benchmark
\end{IEEEkeywords}

\section{Introduction}
The Emergency Events Database recorded 399 natural disasters in 2023, which totalled 86,473 deaths \cite{reliefweb2024}. This was a significant increase from the 387 natural disasters recorded in 2022 totalling 30,704 deaths \cite{reliefweb2023}. Even though a major reason for the massive increase in casualties between the years was due to the earthquakes in Turkey and Syria, the growing severity of natural disasters necessitates enhanced disaster response efforts to prevent large amounts of casualties \cite{reliefweb2024}. 

Quick response times in disasters are essential in fatality prevention, as they allow the potential to rescue many people. For this reason, obtaining Actionable Intelligence — valuable data which is immediately applicable — is crucial \cite{kiatpanont2016extraction}. High quality data is essential because decisions are made during deadly situations \cite{jayawardene2021role}. Remote Sensing has been critical in improving disaster response actions through systematic image extraction from drones and satellites. These extracted images are processed through different techniques, providing emergency managers with essential information for an effective response \cite{merkle2023drones4good}. However, recent research has explored the integration of LLMs in disaster response. 

Large Language Models (LLMs) have shown to have exceptional abilities in many tasks including natural language processing \cite{naveed2023comprehensive}. With more advanced models with improved capabilities such as GPT-4o and Llama 3.1 being developed and released, the future holds promise to integrate LLMs into different fields including disaster response \cite{islam2024gpt}. 

The use of LLMs in disaster response can reduce the time to gather actionable intelligence, and can also serve as a tool for emergency managers when making decisions. It has been shown that fine tuning LLMs with disaster related information allows them to classify disasters based on social media tweets. This improves the ability to quickly respond to disasters, as LLMs can efficiently analyze tweets about a disaster \cite{yin2024crisissense}. In addition, prompting LLMs in specific ways results in them generating reliable action plans comparable to plans a human would develop. This capability can be a valuable tool for emergency managers when determining how to respond in a situation \cite{goecks2023disasterresponsegpt}. 

Despite these innovations, LLMs should only be integrated into disaster response if they are highly accurate in the field. Since disaster response can be life or death, incorrect decisions can result in significant fatalities \cite{myers1994disaster}. To evaluate the abilities of LLMs in disaster response, we propose DisasterQA, the first open source benchmark made for the purpose of gauging the performance of LLMs on disaster response knowledge. In addition, we evaluated five different LLMs along with four different prompting methods on our benchmark. We analyzed performance of each model qualitatively and quantitatively through accuracy and confidence level. By releasing this benchmark, we hope to encourage further evaluating LLMs in disaster response and to encourage the further development of methods to improve their capabilities in this field.

\section{Related Works}

Climate change is only going to continue increasing the occurrences and severity of disasters \cite{o2006climate}. Improved response efficiency can be a means to mitigate these challenges. Recently, large language models (LLMs) have emerged as a tool which can aid in disaster response and relief. 

Goecks and Waytowich developed an algorithm (DisasterResponseGPT) to create a plan of action during a disaster. DisasterResponseGPT utilizes an LLM (GPT-3.5, GPT-4, Bard) that generates several personalized plans for a scenario which the user can request to develop further. The resulting plans of actions were deemed as comparable to human ones, providing an overall increase in efficiency \cite{goecks2023disasterresponsegpt}. 

Additionally, Colverd et al. created a tool, FloodBrain, which produces impact reports from floods with online information. Flood reports, the context and damage of the flood, are often limited by time, but are necessary to begin sending out resources. Colverd et al. utilized LLMs to automate this process, improving flood response times. Various LLMs were tested in the pipeline of FloodBrain, and GPT-4 indicated similar scores on their reports to humans \cite{colverd2023floodbrain}. 

Later on, Otal and Canbaz proposed a framework to identify and classify emergency situations from online social media using Llama2. In addition, they proposed that Llama2 or other LLMs be used to provide instructions to users when the 911 system is overloaded. This framework is valuable, providing a step to integrating LLMs into the 911 system and in delivering instructions to the public during widespread disasters \cite{otal2024llm}. 

Meanwhile, Yin et al introduced a framework to finetune an LLM for multi-label classification of disaster-related tweets. After creating a dataset of disaster related tweets and finetuning an LLM with the data, the resulting model showed higher levels of classification. The LLM was able to classify specific aspects about the disaster, including the type of event and involvement of humanitarian aid, both of which are extremely valuable in a disaster \cite{yin2024crisissense}. 

Furthermore, Fang and Chen examined different prompt engineering techniques to elevate the performance of LLMs in different disasters. After collecting satellite data along with textual data both detailing a disaster, they experimented on four LLMs using three different prompting methods. Later they determined accuracy based on relevance of obtained information \cite{chen2024optimizing}. 

While many have proposed different applications of LLMs in disaster response and relief, there is no set benchmark which can be used to measure an LLMs ability in this field. Making accurate decisions is vital to save as many as possible. We propose our benchmark, made from open source questions online on a variety of topics necessary essential in disaster response. We evaluated five LLMs on our benchmark, along with testing four different prompting methods to see if any resulted in a measurable accuracy increase.

\section{Methods}
\subsection{Dataset Creation}\label{AA}
We spent significant time researching the internet for reliable questions which we could include in our benchmark. We chose to include only multiple choice questions (MCQ) for ease of evaluating the different models’ accuracy. We included questions from six different sources:

\textbf{Quizlet: }We found CEM/AEM practice exam questions on quizlet \cite{quizlet2024}. The International Association of Emergency Managers hosts certification exams for people to become a Certified Emergency Manager (CEM) or Associate Emergency Manager (AEM), both of which are valid certifications \cite{IAEM2024}.                                                                    

\noindent Example Quizlet Question: Frequency, magnitude, potential intensity, and probable spatial extent are all factors to consider when developing a(n):

\noindent Answer Choices:

\noindent A. Hazard profile.

\noindent B. Impact estimate.

\noindent C. Jurisdiction analysis.

\noindent D. Resource assessment.

\noindent Answer: A - Hazard profile

\textbf{Scribd: }We found three documents on Scribd containing MCQs on Disaster Management. We additionally found another document with questions on remote sensing and GIS applications, both of which are vital in disaster response \cite{scribd2024a, scribd2024b, scribd2024c, scribd2024d}
. 

\noindent Example Scribd Question: A sensing system that generates and uses its own energy to illuminate the target is a \_\_\_\_ sensing system:

\noindent Answer Choices:

\noindent A. Passive

\noindent B. Active

\noindent C. radioactive

\noindent D. piezoelectric

\noindent Answer: B - Active

\textbf{Youtube: }We found two different youtube videos going over MCQs on Disaster Management \cite{youtube2024a, youtube2024b}. 

\noindent Example Youtube Question:\textbf{ }Which part of the Disaster Management Cycle does vulnerability analysis come in?:

\noindent Answer Choices:

\noindent A. Response

\noindent B. Preparedness

\noindent C. Mitigation

\noindent D. Recovery

\noindent Answer: C - Mitigation

\textbf{Quizizz:} We found a quiz on Quizizz that tests knowledge of natural disasters \cite{quizizz2024}. This knowledge is vital to maintain situational awareness and expect potential damage during different events. 

\noindent Example Quizizz Question: What causes a tsunami?

\noindent Answer Choices:

\noindent A. Earthquake on land

\noindent B. High Winds

\noindent C. Hurricanes

\noindent D. Earthquake on the seafloor

\noindent Answer: D - Earthquake on the seafloor

\textbf{Slideshare:} We found a document containing Disaster Management MCQs \cite{slideshare2024}.

\noindent Example Slideshare Question: During which phase of the Disaster Management Cycle does an epidemiologist play the most limited role in?

\noindent Answer Choices:

\noindent A. Preparedness

\noindent B. Response

\noindent C. Recovery

\noindent D. Mitigation

\noindent Answer: D - Mitigation

\textbf{Online PDFs:} We found two online PDFs with additional questions on Disaster Management \cite{highlight2024, sdeuoc2024}
.

\noindent Example Online PDFs Question: Most major tsunamis are produced by earthquakes with magnitudes greater than:

\noindent Answer Choices:

\noindent A. 6

\noindent B. 7

\noindent C. 8

\noindent D. 9

\noindent Answer: A - 6

To avoid duplicate questions, before adding any new questions we used the quick search command to check if the same question already existed. The dataset we compiled had 707 questions, offering a decent amount for evaluations. 

\subsection{LLM Prompting}
We chose to test five different LLMs on our benchmark dataset: GPT-3.5 Turbo, GPT-4 Turbo, GPT-4o, Llama 3.1, and Gemini. For Llama 3.1, we used the 8B Instruct version. For Gemini, we used the 1.5 Flash version. Choosing these five models offered a wide coverage for our evaluations.

For ease in extracting each answer produced by the LLMs, we added this phrase at the end of every prompt: “Please answer the question above and end your answer with 'The answer is \{letter\}', where letter is A, B, C, D, E, or F.". Without adding in the phrase, the output generations indicating the LLMs' answer choice exhibits variability. For example, two different outputs generated without the phrase include: "Therefore, the correct answer is: c." and "The best choice among the options provided is c." Once the phrase is added, the output often ends with "The answer is \_",  enabling us to readily obtain the model's selected answer.

To assess whether different prompting methods would improve performances, we evaluated four different prompting methods incorporating self consistency with each. 

\noindent \textbf{No Prompting: }We prompted the model solely with the question and the phrase mentioned above.

\noindent\textbf{Directional Stimulus Prompting: }This method was introduced to guide LLMs into a specific mode of output generation by providing deliberate hints and cues \cite{li2024guiding}. Seeing its potential to improve accuracy and results, we implemented this technique by adding this phrase before each question: “You are an emergency manager. Your responses will dictate survival outcomes for both civilians and emergency responders.” This phrase was intended to guide the LLM to the perspective of an emergency manager during a disaster. 

\noindent\textbf{Chain of Thought Prompting: }Chain of Thought prompting has been shown to enhance the performance of LLMs on reasoning tasks \cite{wei2022chain}. We utilized zero-shot Chain of Thought prompting by not providing examples, but instead added the phrase: “Let’s think step by step.” before each question to encourage reasoning in the LLMs when generating a response. 

\noindent\textbf{Few Shot Prompting: }LLMs are able to learn from examples, and few shot prompting leverages this allowing in context learning \cite{ma2023fairness}. We applied few shot prompting with three question answer examples related to disaster management. We added this phrase before each question: 

\noindent “\textbf{Question:} 
Effective delegation involves \_\_\_\_\_\_\_\_.

\noindent \textbf{Answer Choices:}

\noindent A. Determining that all credit goes to the team leader.

\noindent B. Ensuring that all tasks are equally distributed.

\noindent C. Ensuring that the person has the necessary authority to do the job properly.

\noindent D. Guaranteeing all employees have a chance to do the same task.

\noindent \textbf{Answer:} Effective delegation involves assigning tasks and responsibilities to others while ensuring that the person has the necessary authority to do the job properly. This includes providing clear instructions, setting expectations, and offering support as needed. The answer is C.

\noindent \textbf{Question: }What should a leader do when setting up implementation structures to facilitate change?

\noindent \textbf{Answer Choices:}

\noindent A. Focus on interim systems because there will be time later to deal with long-term impact.

\noindent B. Consider primarily long-term impacts because the short-term will take care of itself.

\noindent C. Include representatives of all key stakeholder groups.

\noindent D. Place the greatest emphasis on how the change will impact the highest levels of management.

\noindent \textbf{Answer:} When setting up implementation structures to facilitate change, a leader should include representatives of all key stakeholder groups. This ensures that the perspectives and needs of all parties involved are considered, leading to a more successful implementation of change. The answer is C.

\noindent \textbf{Question:} An incident command system provides for a common organizational structure that enables effective and efficient incident management. It is designed to ensure which of the following?

\noindent \textbf{Answer Choices:}

\noindent A. Achievement of the entity's strategic objectives.

\noindent B. Efficient use of resources.

\noindent C. Health and safety of the Whole Community.

\noindent D. Use of specific agency or organizational codes and acronyms.

\noindent \textbf{Answer:} An incident command system provides for a common organizational structure that enables effective and efficient incident management. It is designed to ensure the achievement of the entity's strategic objectives, efficient use of resources, and health and safety of the Whole Community. The answer is A.”

\noindent \textbf{Emotional Prompting: }Prompting LLMs with emotion has been shown to yield better results. Just as human responses can be influenced by emotional context, LLMs may generate better responses after including emotional cues \cite{li2023large}. To illustrate the dire situation and emphasize the need for correct answers, we preceded each question with this phrase: “A disaster is happening, so your answer to this question is of the utmost importance. People may die given an incorrect answer. Only give an answer once you are as confident as you can be as the fate of some humans’ livelihoods rests on your hands.”

\noindent\textbf{Self Consistency: }Self Consistency involves generating many responses from the LLM and selecting the most common answer. This approach has been shown to improve performance on reasoning tasks \cite{wang2022self}. We implemented Self Consistency by prompting each question five times and recording each of the generations. We would later compare first index accuracies with majority vote (Self Consistency) accuracies. 

\section{Results}
We extracted the answer from each generation using a regular expression which identifies the first letter after any of the following phrases: “answer is,” “answer here is,” “answer choice is,” “answer choice here is,” “choices provided is:,” “options provided is:,” “from the list is,” “choices is,” “one out is.” We included many variations, as shown from these phrases, informed from previous model outputs, to maximize the chance of obtaining answer choices to as many questions as possible. 

We generated five completions for each question. The first index accuracy was calculated based on the answer choice from the initial completion, while the majority vote accuracy was determined by identifying the most frequent answer choice across all five completions. Some models tended to not answer certain questions. To account for this factor, the number of unanswered questions was subtracted from the denominator in accuracy calculations. 
\begin{table}[h]
    \centering
    \begin{tabular}{|c|c|}
        \hline
        Model & Avg number of questions \\ \hline
        GPT-3.5 Turbo & 3.8 \\ \hline
        GPT-4 Turbo & 2.8 \\ \hline
        GPT-4o & 3 \\ \hline
        Gemini 1.5 Flash & 37.6 \\ \hline
        Llama 3.1-8B Instruct & 5.8 \\ \hline
    \end{tabular}
    \caption{Average Number of Unanswered Questions per model for all prompting methods}
    \label{tab:unanswered_questions}
\end{table}

Gemini had a significantly higher number of unanswered questions compared to other models. This difference is partly because of Gemini blocking certain questions due to safety guidelines. 
Example of a question from our dataset which was blocked by Gemini:

\noindent \textbf{Question:} What is a lahar?

\noindent \textbf{Answer Choices:} 

\noindent A. Magma 

\noindent B. Lava 

\noindent C. Volcanic mud flow 

\noindent D. None of the above

\noindent \textbf{Completion:} None

Furthermore, when using our emotional prompt, Gemini failed to answer 73 questions, which is 10.3\% of our dataset. 
Example of a question from our dataset which Gemini did not answer during emotional prompt:

\noindent \textbf{Question:} The total amount of water requirement per liter for single cholera patient per day is

\noindent \textbf{Answer Choices:} 

\noindent  A. 80 

\noindent B. 60 

\noindent C. 40 

\noindent D. 100 

\noindent E. 50

\noindent \textbf{Completion:} “I understand the urgency of the situation and the potential consequences of an incorrect answer. However, I am an AI language model and am not equipped to provide medical advice. 
**Providing a specific fluid volume for a cholera patient is extremely dangerous and could lead to serious complications.**  Fluid requirements for cholera patients are highly individualized and depend on factors like:
* **Severity of dehydration:**  Patients with severe dehydration need much more fluid than those with mild dehydration.
* **Age:** Children and older adults require different fluid volumes than adults.
* **Overall health status:** Pre-existing conditions can affect fluid needs.
**It is crucial to consult a qualified medical professional for accurate treatment and fluid management.** 
Please seek immediate medical attention for any cholera patients. 
**Do not rely on AI for medical advice.**”

This is an issue with Gemini, and the accuracies we recorded may be misleading as we did not impact a model’s accuracy for not answering a question. From the emotional prompting completion, Gemini seems to not respond because it believes it is not a reliable source. The pressure added from our emotional prompt contributes to this, indicating that emotional prompting hinders Gemini’s willingness to answer difficult questions. Other models tended answer far more questions than Gemini. Surprisingly, GPT-4 Turbo on average answered the most questions, even outperforming GPT-4o in terms of fewer unanswered questions. Occasionally, models answered with “none" when that was not an option or simply restated the question without an answer.  

Along with accuracy, we also recorded log probabilities (Logprobs) for further analysis. We used the Logprobs to calculate the average confidence each model had in its answer choice for each completion. Since we instructed the model to end its completion with its answer choice followed by a period, we analyzed the Logprobs of the second last tokens. For each question, we converted the first Logprob of the second last token into confidence for all five completions. We then calculated the average confidence from all of these second last Logprob confidence values. Gemini API unfortunately does not support Logprob storage, but we did this analysis for the other models. 
\begin{center}
    \textit{* The best performance for each LLM is bolded}
\end{center}
\vspace{-10pt} 

\begin{table}[H]
    \centering
    \captionsetup{justification=raggedright, singlelinecheck=false} 
    \begin{tabular}{|c|c|c|c|}
        \hline
        Model & 1st Index & Majority Vote & Avg Confidence \\ \hline
        GPT-3.5 Turbo & \textbf{74.72\%} & \textbf{74.72\%} & 75.69\% \\ \hline
        GPT-4 Turbo & 81.70\% & 82.10\% & 38.76\% \\ \hline
        GPT-4o & 82.84\% & 84.09\% & 79.09\% \\ \hline
        Gemini 1.5 Flash & 79.64\% & 78.97\% & N/A \\ \hline
        Llama 3.1-8B Instruct & 73.29\% & 73.10\% & 97.40\% \\ \hline
    \end{tabular}
    \caption{Results from No Prompting}
    \label{tab:four_by_five_table}
\end{table}
\vspace{-0.36em} 
\begin{table}[H]
    \centering
    \captionsetup{justification=raggedright, singlelinecheck=false} 
    \begin{tabular}{|c|c|c|c|}
        \hline
        Model & 1st Index & Majority Vote & Avg Confidence \\ \hline
        GPT-3.5 Turbo & 72.87\% & 73.40\% & 69.67\% \\ \hline
        GPT-4 Turbo & 81.11\% & 82.13\% & 38.90\% \\ \hline
        GPT-4o & 82.98\% & 84.11\% & 76.22\% \\ \hline
        Gemini 1.5 Flash & 78.07\% & 80.09\% & N/A \\ \hline
        Llama 3.1-8B Instruct & 71.12\% & 71.02\% & \textbf{98.59\%} \\ \hline
    \end{tabular}
    \caption{Results from Directional Stimulus Prompting}
    \label{tab:directional_stimulus_prompting_table}
\end{table}

\begin{table}[H]
    \centering
    \captionsetup{justification=raggedright, singlelinecheck=false} 
    \begin{tabular}{|c|c|c|c|}
        \hline
        Model & 1st Index & Majority Vote & Avg Confidence \\ \hline
        GPT-3.5 Turbo & 72.61\% & 72.97\% & 78.17\% \\ \hline
        GPT-4 Turbo & 81.28\% & 82.13\% & \textbf{70.56\%} \\ \hline
        GPT-4o & \textbf{84.21\%} & \textbf{85.78\%} & 77.02\% \\ \hline
        Gemini 1.5 Flash & 78.73\% & 80.56\% & N/A \\ \hline
        Llama 3.1-8B Instruct & 72.21\% & 72.74\% & 96.87\% \\ \hline
    \end{tabular}
    \caption{Results from Chain of Thought Prompting}
    \label{tab:chain_of_thought_prompting_table}
\end{table}

\begin{table}[H]
    \centering
    \captionsetup{justification=raggedright, singlelinecheck=false} 
    \begin{tabular}{|c|c|c|c|}
        \hline
        Model & 1st Index & Majority Vote & Avg Confidence \\ \hline
        GPT-3.5 Turbo & 73.04\% & 73.12\% & 70.07\% \\ \hline
        GPT-4 Turbo & \textbf{83.19\%} & \textbf{83.21\%} & 21.83\% \\ \hline
        GPT-4o & 83.64\% & 84.92\% & 51.91\% \\ \hline
        Gemini 1.5 Flash & 78.31\% & 79.73\% & N/A \\ \hline
        Llama 3.1-8B Instruct & \textbf{73.35\%} & \textbf{73.53\%} & 98.22\% \\ \hline
    \end{tabular}
    \caption{Results from Few Shot Prompting}
    \label{tab:few_shot_prompting_table}
\end{table}

\begin{table}[H]
    \centering
    \captionsetup{justification=raggedright, singlelinecheck=false} 
    \begin{tabular}{|c|c|c|c|}
        \hline
        Model & 1st Index & Majority Vote & Avg Confidence \\ \hline
        GPT-3.5 Turbo & 71.45\% & 73.83\% & \textbf{79.21\%} \\ \hline
        GPT-4 Turbo & 80.40\% & 80.82\% & 47.38\% \\ \hline
        GPT-4o & 83.97\% & 85.11\% & \textbf{81.44\%} \\ \hline
        Gemini 1.5 Flash & \textbf{81.21\%} & \textbf{82.02\%} & N/A \\ \hline
        Llama 3.1-8B Instruct & 70.50\% & 69.93\% & 95.57\% \\ \hline
    \end{tabular}
    \caption{Results from Emotional Prompting}
    \label{tab:emotional_prompting_table}
\end{table}

\section{Analysis}
Prompting seems to have not made a large difference in accuracy, which generally falls around the same range across different prompting methods. Additionally, different models performed better with different prompting methods. For example, GPT-3.5 Turbo performed best with no prompting while GPT-4o performed best with Chain of Thought Prompting. There was an increase in accuracy from GPT-3.5 Turbo to GPT-4 Turbo to GPT-4o. Llama’s accuracy was around the same or slightly worse than GPT-3.5 Turbo for the different prompting methods. Gemini’s accuracy was close to and sometimes slightly lower than GPT-4 Turbo’s accuracy. However we did not penalize unanswered questions, which could have affected Gemini’s relative performance. Using a two-sample proportion Z-test, the accuracy differences across prompting methods were not statistically significant at the 0.05 p-value threshold for any of the models. Additionally, the majority vote accuracy is consistently higher or around the same as the first index accuracy. 

Llama consistently exhibited the highest average confidence in its answer choice for all the prompting methods, despite consistently having one of the lowest accuracies on the benchmark. GPT-4o exhibited high confidence but had a sharp decline in confidence with Few Shot Prompting. GPT-3.5 Turbo consistently exhibited high confidence, with the highest being with emotional prompting. GPT-4 Turbo exhibited the lowest confidence by a large margin, however its confidence significantly increased during Chain of Thought Prompting.

In general, GPT-4o tends to consider all of the answer choices in its generations. It generates a detailed decision often having the longest answers. Gemini typically generates long responses and sometimes considers all of the answer choices, but not as consistently as GPT-4o. GPT-4 Turbo also tends to consider all of the answer choices, but to a lesser extent than GPT4o. Llama exhibits the same behavior, although its responses tend to be shorter than those two models. GPT-3.5 Turbo however rarely considers all answer choices, often providing the shortest answers out of all the models.

We also analyzed questions the models consistently answered incorrectly. Specifically, we compiled a list of questions in our dataset where every model’s majority vote was incorrect under Directional Stimulus Prompting: [22, 28, 30, 43, 46, 54, 87, 90, 91, 97, 100, 102, 115, 127, 131, 134, 136, 142, 153, 154, 217, 256, 263, 267, 276, 289, 305, 347, 367, 374, 396, 401, 415, 429, 437, 444, 445, 522, 549, 558, 563, 580, 581, 591, 601, 613, 618, 623, 640, 645, 646, 656, 658, 671, 675, 694, 700]. After analyzing the questions, we observed that the models tended to get questions wrong from various areas in disaster response, not a specific area. These areas included emergency management knowledge, disaster knowledge, and GIS/remote sensing applications. 

\noindent \textbf{Example Question:} What activity takes place during the response phase of the disaster cycle?

\noindent \textbf{Answer Choices:}

\noindent A. Conducting surveillance of health problems

\noindent B. Conducting an inventory of available resources

\noindent C. Training of health personnel

\noindent D. Conducting epidemiological studies

\noindent \textbf{Answer:} A - Conducting surveillance of health problems

\noindent GPT-3.5 Turbo answered B

\noindent GPT-4 Turbo answered A and B, max vote was B

\noindent GPT-4o answered B

\noindent Gemini answered B

\noindent Llama 3.1 Instruct answered B

\noindent \textbf{Example Question:} During volcanic eruption, the extreme risk zone is within a distance of:

\noindent \textbf{Answer Choices:}

\noindent A. Up to 100 meters

\noindent B. 100 to 300 meters

\noindent C. 300 meters to 3 kilometers

\noindent D. None of the above

\noindent \textbf{Answer:} A - Up to 100 meters

\noindent GPT-3.5 Turbo answered C

\noindent GPT-4 Turbo answered D

\noindent GPT-4o answered D and C

\noindent Gemini answered D

\noindent Llama 3.1 Instruct answered C

\noindent \textbf{Example Question:} Effects on radiance in images caused due to variations in topography can be mitigated by:

\noindent \textbf{Answer Choices:}

\noindent a. Enhancement

\noindent b. Band ratioing

\noindent c. Image fusion

\noindent d. Radiometric correction

\noindent \textbf{Answer:} B - Band Ratioing

\noindent GPT-3.5 Turbo answered D

\noindent GPT-4 Turbo answered D

\noindent GPT-4o answered D

\noindent Gemini answered D

\noindent Llama 3.1 Instruct answered D

The models somewhat tend to consistently select the same incorrect answer choice, as shown from these examples and other questions we analyzed. This pattern indicates a potential knowledge issue with all LLMs, rather than just a specific model. The overall findings emphasize the need to further advance LLMs in disaster response knowledge, as accuracy on benchmarks should approach 100\% for reliability in critical situations.

\section{Limitations}
Our benchmark consists of 707 questions, which is a substantial amount, but it could be expanded. Evaluating a larger set of questions would offer a better representation of an LLM's capabilities in disaster response. A larger set of questions may also cover a broader range of questions and topics we may have missed in our benchmark. Additionally, we did not account for unanswered questions, if we did the results may have been different, especially for Gemini. Some models' answer choices were embedded within the completion and were not extracted by our regular expression function. Due to time constraints, we could not manually examine cases like these. 

\section{Conclusion}
LLMs have shown advanced capabilities in many fields, including disaster response. However, the absence of a standardized benchmark for evaluating LLMs in disaster response pushed us to develop DisasterQA, an open source benchmark with 707 questions covering various topics on disaster response. We created this dataset from 6 online sources, ensuring no duplicate questions.

We evaluated five LLMs: GPT-3.5 Turbo, GPT-4 Turbo, GPT-4o, Gemini 1.5 Flash, and Llama 3.1-8B Instruct on our benchmark and applied four different prompting methods to each model. GPT-4o obtained the highest accuracy of 85.78\% on our dataset with Chain of Thought prompting. Regardless of each prompting method, each models accuracy was around the same. 

We also analyzed the confidence each model had in its answer choice by storing the Logprobs. Llama had the highest confidence, nearly 100\% despite having one of the lowest accuracies. Additionally, some models tended to not answer questions, especially Gemini, which had an average of 37.6 unanswered questions over all the prompting methods. Although we did not account for this factor in our accuracy calculations, it is something to consider when evaluating a model's effectiveness. 

After analyzing questions most models got incorrect, we concluded that there was not a specific topic that the models were lacking knowledge on. They were incorrect on questions from a wide range of disaster response topics including GIS applications and natural disaster knowledge. 

We believe LLMs need to be further improved, with accuracies approaching 100\% in the field of disaster response. We hope that the release of our benchmark to HuggingFace (DisasterQA) encourages further fine-tuning and development of LLMs in disaster response. With improvements in accuracy, LLMs could become valuable tools providing advice to emergency managers and other disaster response personnel, and this benchmark is a step toward that goal. 

\section*{Acknowledgment}
We would like to thank Sam Scheele for helping refine the research idea. We would also want to thank Kevin Zhu and Algoverse AI for useful feedback on the paper and providing credits to run the LLM generations.


\begin{thebibliography}{10}

\bibitem{reliefweb2024}
``2023 disasters in numbers,'' tech. rep., ReliefWeb, 2024.
\newblock Accessed: August 10, 2024.

\bibitem{reliefweb2023}
``2022 disasters in numbers,'' tech. rep., ReliefWeb, 2023.
\newblock Accessed: August 10, 2024.

\bibitem{kiatpanont2016extraction}
R.~Kiatpanont, U.~Tanlamai, P.~Chongstitvatana, {\em et~al.}, ``Extraction of actionable information from crowdsourced disaster data,'' {\em Journal of emergency management}, vol.~14, no.~6, pp.~377--390, 2016.

\bibitem{jayawardene2021role}
V.~Jayawardene, T.~J. Huggins, R.~Prasanna, and B.~Fakhruddin, ``The role of data and information quality during disaster response decision-making,'' {\em Progress in disaster science}, vol.~12, p.~100202, 2021.

\bibitem{merkle2023drones4good}
N.~Merkle, R.~Bahmanyar, C.~Henry, S.~M. Azimi, X.~Yuan, S.~Schopferer, V.~Gstaiger, S.~Auer, A.~Schneibel, M.~Wieland, {\em et~al.}, ``Drones4good: Supporting disaster relief through remote sensing and ai,'' in {\em Proceedings of the IEEE/CVF International Conference on Computer Vision}, pp.~3770--3774, 2023.

\bibitem{naveed2023comprehensive}
H.~Naveed, A.~U. Khan, S.~Qiu, M.~Saqib, S.~Anwar, M.~Usman, N.~Akhtar, N.~Barnes, and A.~Mian, ``A comprehensive overview of large language models,'' {\em arXiv preprint arXiv:2307.06435}, 2023.

\bibitem{islam2024gpt}
R.~Islam and O.~M. Moushi, ``Gpt-4o: The cutting-edge advancement in multimodal llm,'' {\em Authorea Preprints}, 2024.

\bibitem{yin2024crisissense}
K.~Yin, C.~Liu, A.~Mostafavi, and X.~Hu, ``Crisissense-llm: Instruction fine-tuned large language model for multi-label social media text classification in disaster informatics,'' {\em arXiv preprint arXiv:2406.15477}, 2024.

\bibitem{goecks2023disasterresponsegpt}
V.~G. Goecks and N.~R. Waytowich, ``Disasterresponsegpt: Large language models for accelerated plan of action development in disaster response scenarios,'' {\em arXiv preprint arXiv:2306.17271}, 2023.

\bibitem{myers1994disaster}
D.~Myers, ``Disaster response and recovery,'' {\em US Department of Health and Human Services, Rockville MD}, 1994.

\bibitem{o2006climate}
G.~O'Brien, P.~O'keefe, J.~Rose, and B.~Wisner, ``Climate change and disaster management,'' {\em Disasters}, vol.~30, no.~1, pp.~64--80, 2006.

\bibitem{colverd2023floodbrain}
G.~Colverd, P.~Darm, L.~Silverberg, and N.~Kasmanoff, ``Floodbrain: Flood disaster reporting by web-based retrieval augmented generation with an llm,'' {\em arXiv preprint arXiv:2311.02597}, 2023.

\bibitem{otal2024llm}
H.~T. Otal, E.~Stern, and M.~A. Canbaz, ``Llm-assisted crisis management: Building advanced llm platforms for effective emergency response and public collaboration,'' in {\em 2024 IEEE Conference on Artificial Intelligence (CAI)}, pp.~851--859, IEEE, 2024.

\bibitem{chen2024optimizing}
W.~Chen and J.~Fang, ``Optimizing ai-driven disaster management through llms,'' 2024.

\bibitem{quizlet2024}
Quizlet, ``Iaem cem/aem practice exam.'' \url{https://quizlet.com/710141954/iaem-cemaem-practice-exam-flash-cards/}, 2022.
\newblock Accessed: July 15, 2024.

\bibitem{IAEM2024}
{International Association of Emergency Managers}, ``Iaem official website.'' \url{https://www.iaem.org/}.
\newblock Accessed: August 11, 2024.

\bibitem{scribd2024a}
Scribd, ``Disaster management multiple-choice question with answers.'' \url{https://www.scribd.com/document/498544241/Disaster-Management-Multiple-choice-question-with-answers}, 2021.
\newblock Accessed: July 15, 2024.

\bibitem{scribd2024b}
Scribd, ``Disaster management mcqs.'' \url{https://www.scribd.com/document/511504220/Disaster-Management-MCQs}, 2021.
\newblock Accessed: July 15, 2024.

\bibitem{scribd2024c}
Scribd, ``S-skill disaster management qa.'' \url{https://www.scribd.com/document/491235815/S-Skill-Disaster-Management-QA}, 2021.
\newblock Accessed: July 15, 2024.

\bibitem{scribd2024d}
Scribd, ``Gis master mcq merged.'' \url{https://www.scribd.com/document/515895039/GIS-Master-MCQ-merged}, 2021.
\newblock Accessed: July 16, 2024.

\bibitem{youtube2024a}
``100 important mcqs disaster management. entrance exam and all other exams..'' \url{https://www.youtube.com/watch?v=Jp\_Skd\_YSHc&t=31s}, 2022.
\newblock Accessed: July 14, 2024.

\bibitem{youtube2024b}
``Disaster management mcq | ge8071 mcq | dm mcq | 100 important questions.'' \url{https://www.youtube.com/watch?v=OzU68Ko3R-k}, 2020.
\newblock Accessed: July 17, 2024.

\bibitem{quizizz2024}
``Natural disaster.'' \url{https://quizizz.com/admin/quiz/58dbde3cfed1bc635b943a6a/natural-disaster}, 2017.
\newblock Accessed: July 18, 2024.

\bibitem{slideshare2024}
``Mcq in disaster management.'' \url{https://www.slideshare.net/slideshow/mcq-in-diaster-magement/250334432}, 2021.
\newblock Accessed: July 15, 2024.

\bibitem{highlight2024}
``300+ elements of remote sensing multiple choice questions answers.'' \url{https://www.highlightcomputer.com/ElementsofRemoteSensing.pdf}, 2020.
\newblock Accessed: July 19, 2024.

\bibitem{sdeuoc2024}
``Question bank of 2 semester degree audit course - 2e02 : Disister management.'' \url{https://sde.uoc.ac.in/sites/default/files/sde_videos/Disaster%20Managment%20%20MCQ-2-36.pdf}.
\newblock Accessed: July 21, 2024.

\bibitem{li2024guiding}
Z.~Li, B.~Peng, P.~He, M.~Galley, J.~Gao, and X.~Yan, ``Guiding large language models via directional stimulus prompting,'' {\em Advances in Neural Information Processing Systems}, vol.~36, 2024.

\bibitem{wei2022chain}
J.~Wei, X.~Wang, D.~Schuurmans, M.~Bosma, F.~Xia, E.~Chi, Q.~V. Le, D.~Zhou, {\em et~al.}, ``Chain-of-thought prompting elicits reasoning in large language models,'' {\em Advances in neural information processing systems}, vol.~35, pp.~24824--24837, 2022.

\bibitem{ma2023fairness}
H.~Ma, C.~Zhang, Y.~Bian, L.~Liu, Z.~Zhang, P.~Zhao, S.~Zhang, H.~Fu, Q.~Hu, and B.~Wu, ``Fairness-guided few-shot prompting for large language models,'' {\em Advances in Neural Information Processing Systems}, vol.~36, pp.~43136--43155, 2023.

\bibitem{li2023large}
C.~Li, J.~Wang, Y.~Zhang, K.~Zhu, W.~Hou, J.~Lian, F.~Luo, Q.~Yang, and X.~Xie, ``Large language models understand and can be enhanced by emotional stimuli,'' {\em arXiv preprint arXiv:2307.11760}, 2023.

\bibitem{wang2022self}
X.~Wang, J.~Wei, D.~Schuurmans, Q.~Le, E.~Chi, S.~Narang, A.~Chowdhery, and D.~Zhou, ``Self-consistency improves chain of thought reasoning in language models,'' {\em arXiv preprint arXiv:2203.11171}, 2022.

\end{thebibliography}
\end{document}